\documentclass[10pt,twocolumn,letterpaper]{article}

\usepackage{cvpr}
\usepackage{times}
\usepackage{epsfig}
\usepackage{graphicx}
\usepackage{amsmath}
\usepackage{amssymb}

\usepackage{lipsum}

\usepackage[pagebackref=true,breaklinks=true,letterpaper=true,colorlinks,bookmarks=false]{hyperref}

\cvprfinalcopy % *** Uncomment this line for the final submission

\def\cvprPaperID{*}%10425} % *** Enter the CVPR Paper ID here

\def\Ttheta{\boldsymbol\theta}
\def\bb{\mathbf b}

\def\TD{\mathbf D}
\def\TK{\mathbf K}

\def\TY{\mathbf Y}

\def\TW{\mathbf W}

\newcommand{\bfK}{{\bf K}}

\newcommand{\bfW}{{\bf W}}

\newcommand{\bfY}{{\bf Y}}

\newcommand{\bfb}{{\bf b}}

\newcommand{\bftheta}{{\boldsymbol \theta}}

\ifcvprfinal\fi
\begin{document}

%%%%%%%%% TITLE
\title{Symmetric block-low-rank layers for fully reversible multilevel neural networks}

\author{Bas Peters\\
Computational Geosciences Inc.\\
1623 West 2nd Ave, Vancouver, BC, Canada\\
{\tt\small bas@compgeoinc.com}
% For a paper whose authors are all at the same institution,
% omit the following lines up until the closing ``}''.
% Additional authors and addresses can be added with ``\and'',
% just like the second author.
% To save space, use either the email address or home page, not both
\and
Eldad Haber\\
University of British Columbia\\
2207 Main Mall, Vancouver, BC, Canada\\
{\tt\small ehaber@eoas.ubc.ca}
\and
Keegan Lensink\\
University of British Columbia\\
2207 Main Mall, Vancouver, BC, Canada\\
{\tt\small klensink@eoas.ubc.ca}
}

\maketitle
%\thispagestyle{empty}

%%%%%%%%% ABSTRACT
\begin{abstract}
Factors that limit the size of the input and output of a neural network include memory requirements for the network states/activations to compute gradients, as well as memory for the convolutional kernels or other weights. The memory restriction is especially limiting for applications where we want to learn how to map volumetric data to the desired output, such as video-to-video. Recently developed fully reversible neural networks enable gradient computations using storage of the network states for a couple of layers only. While this saves a tremendous amount of memory, it is the convolutional kernels that take up most memory if fully reversible networks contain multiple invertible pooling/coarsening layers. Invertible coarsening operators such as the orthogonal wavelet transform cause the number of channels to grow explosively. We address this issue by combining fully reversible networks with layers that contain the convolutional kernels in a compressed form directly. Specifically, we introduce a layer that has a symmetric block-low-rank structure. In spirit, this layer is similar to bottleneck and squeeze-and-expand structures. We contribute symmetry by construction, and a combination of notation and flattening of tensors allows us to interpret these network structures in linear algebraic fashion as a block-low-rank matrix in factorized form and observe various properties. A video segmentation example shows that we can train a network to segment the entire video in one go, which would not be possible, in terms of memory requirements, using non-reversible networks and previously proposed reversible networks. 
\end{abstract}

%%%%%%%%% BODY TEXT
\section{Introduction}

We consider memory limitations associated with neural-networks that map 3D data to 3D output for applications, such as semantic segmentation of video. Networks with a limited number of layers and small ($3 \times 3 \times 3$) convolutional kernels can still learn from large scale structure/information in the input data, if we employ a multi-resolution or multi-level network that increases the effective field-of-view, or, receptive field. Networks in this category include U-nets \cite{Ronneberger_2015} and various encoder-decoder type networks.

The dominant factor that limits the input data size and network depth is the storage of the network state at each layer to compute a gradient of the loss function using back-propagation, often implemented via reverse-mode automatic differentiation. Recomputation of the network (forward) states in reverse order during back-propagation avoids this problem. This recomputation is possible using fully reversible hyperbolic networks \cite{lensink2019fully} for image/video segmentation. That work extends fully reversible networks for image classification \cite{jacobsen2018irevnet,leemput2018memcnn} and networks that are reversible in between coarsening/pooling stages only \cite{Chang2017Reversible,GomezEtAl2017,DinhSB16}. 

\begin{figure*}
\begin{center}
   \includegraphics[width=1.0\linewidth]
                   {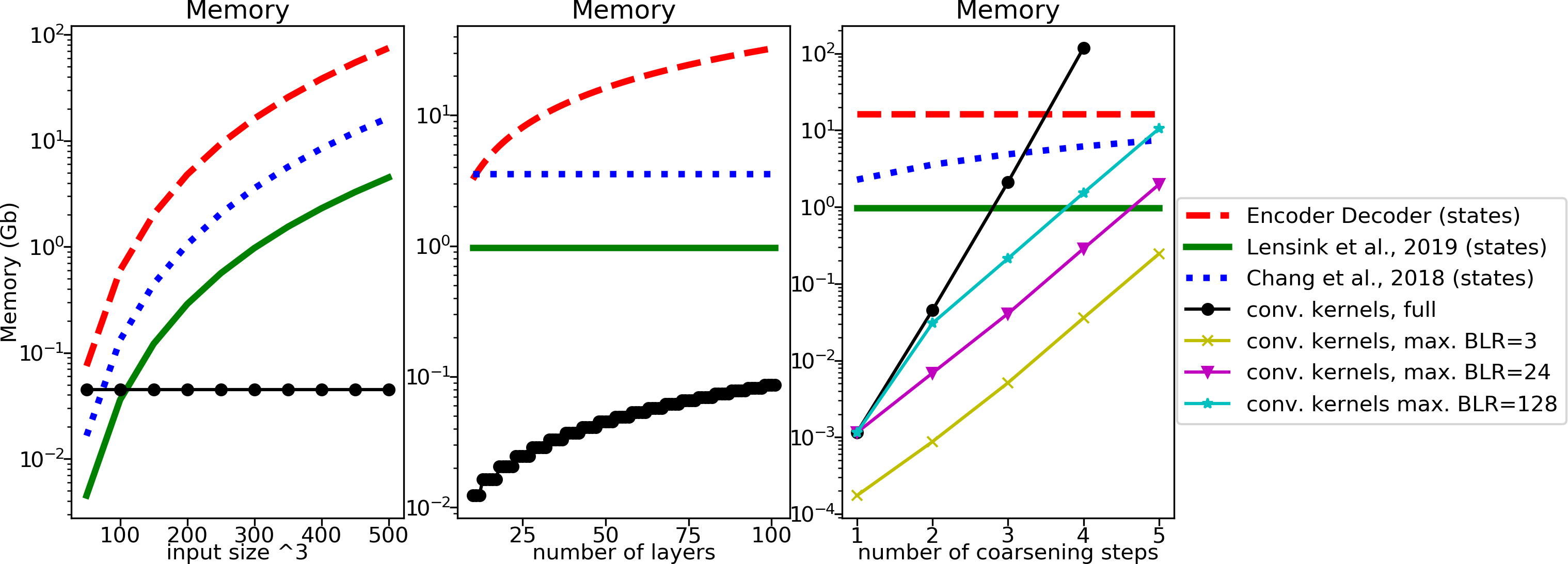}
\end{center}
   \caption{Memory requirements (Gigabyte) for network states (activations) and convolutional kernels. Left: as a function of input size and a fixed $50$ layer network with two coarsening stages. Middle: as a function of an increasing number of layers but with fixed input size ($300^3$) and fixed number of two coarsenings. Right: as a function of an increasing number of coarsening steps but with fixed number of layers ($50$) and input size ($300^3$). Our proposed Block Low-Rank (BLR) layers avoid an exploding number of convolutional kernels with increased coarsening in reversible networks.}
\label{fig:memorycurves}
\end{figure*}

Fully reversible networks have a constant memory requirement that is independent of network depth, see Figure \ref{fig:memorycurves}. In that case, the convolutional kernels become the dominant memory consumer if we coarsen the data several times in the network. Whereas networks based on the ResNet layer \cite{he2016deep} often compress, i.e., the number of channels does not increase by a factor eight when we coarsen in 3D by a factor two in each of the three directions, we need to increase the number of channels by a factor eight when we coarsen in a fully reversible network. This preservation of the number of elements in the tensors is used to make the coarsening and channel-count changes invertible operations. \cite{lensink2019fully} use the Haar transform and \cite{DinhSB16,jacobsen2018irevnet} reorganize the data via checkerboard ordering into a tensor with multiple channels and decreased resolution. Coarsening while preserving the number of tensor elements leads to an `explosion' of the number of channels. For instance, if the input is three-channel RGB, there are $192$ channels after two coarsening layers, and an astonishing $98304$ channels in case we wish to coarsen five times. The storage and computations of the associated $98304^2$ convolutional kernels (just for one layer at the coarsest level) would be completely unfeasible. See Figure \ref{fig:memorycurves} for an illustration of this effect.

In this work, we present solutions to the above problems by combining the design of fully reversible networks with layers that reduce the storage and computations related to the convolutional kernels. Network compression methods such as pruning and low-rank \cite{NIPS2014_5544} matrix/tensor factorization train a 'full' network first, followed by compression. These ideas do not apply to our problem because we can never train the full network. We need to employ methods that directly train a reduced-memory network. Work in this field includes replacing most of the convolutional kernels by scalars \cite{ephrath2019leanresnet}, and equipping the convolutional kernels with a block-circulant structure for weight compression \cite{8686678,treister2018low}. Note that there are other techniques in machine learning that also train a model in compressed form directly. For example, LR-factorizations for matrix completion \cite{rennie2005fast,doi:10.1137/130919210}.
 
Inspired by fully reversible networks, network compression, and low-memory network layers, we construct layers that are a low-rank factorization of the convolutional kernels. This structure enables explicit limitations on the number of convolutions in a layer while preserving a possibly extremely large number of channels. Our approach is similar to `squeeze-and-expand' bottleneck methods like \cite{Szegedy_2015_CVPR,he2016deep,iandola2016squeezenet} proposed for residual networks, mainly in the context of image classification. Our method differs in the sense that we have one instead of two non-linearities per layer, and our convolutional kernels have a symmetric positive-semidefinite block-structure, which squeezenet does not have. Also different, is that we do not rely on $1 \times 1 \times 1$ convolutions for compression and expansion. Because we explicitly induce and recognize the block low-rank structure in our network layer, we can directly read of various mathematical properties of interest and show these properties are preserved under the action of pointwise non-negative activation functions.

The second contribution of this work is an adaptive training strategy. We show that if training starts with a block rank that is too low, we can increase the rank while training and still obtain good results. This procedure reduces the number of convolutions that are computed per iteration.

Finally, we note that the proposed layer fits in almost any neural network, but we specifically target problems with large-scale inputs for which we need to use fully reversible networks in combination with the symmetric block-low-rank layer. Numerical experiments verify we can use these tools to solve problems that would not be possible with existing network architectures..

\section{Contributions}
\label{sec1b}

\begin{itemize}

\item We propose a symmetric block-low-rank layer for neural networks that can naturally, and without the need to change code, software, and implementation, induce a block low-rank structure to the convolutional kernels. We emphasize the description `naturally' because the block low-rank structure appears directly in a factorized form, avoiding the need for additional computations to find low-rank decompositions of any kind.

\item We describe when symmetric block-low-rank layers are beneficial compared to regular symmetric layers and `standard' layers fully reversible networks. These benefits appear in applications like semantic segmentation of 3D video, where the network contains multiple coarsening/pooling layers. This work is the first where we combine full reversibility and network compression techniques. As a result, the required memory for network states and convolutional kernels is small and controllable, such that we can use larger data inputs. 

\item We train networks in block low-rank compressed form directly. A lower block-rank of the convolutional kernels does not limit the number of channels. These properties allow us to change the block-rank adaptively while training, either to prevent over-fitting or to increase training data-fitting capabilities.

\item The combination of: a) the structure of the proposed symmetric block-low-rank layer and b) the matrix-vector product notation where we flatten 5-D tensors into block-matrices, allows us to gain new insights about bottleneck/squeeze-expand type layers. We can directly attribute a few properties to these compression-based layers. Such connections do not appear in previously published work.

\end{itemize}

\section{Hyperbolic and fully reversible networks}
Linear algebraic matrix-vector product notation is required for the interpretation that will follow in the following section. Computational implementation can still use 5D tensor format. We flatten tensors that contain network states $\mathbb{R}^{n_x \times n_y \times n_z \times n_\text{chan} }$ to block-vectors $\mathbb{R}^{n_x n_y n_z n_\text{chan}}$ that contain $n_\text{chan}$ sub-vectors,
\begin{equation}
\TY \equiv
\begin{bmatrix}
Y^1 \\
Y^2 \\
\vdots \\
Y^{n_\text{chan}}
\end{bmatrix}.
\end{equation}
 
Similarly, we rewrite the convolutional kernels for a given layer in tensor format $\mathbb{R}^{k_x \times k_y \times k_z \times n_\text{chan out} \times n_\text{chan in}}$ as the block-matrix $\TK$ with $n_\text{chan out}$ rows and $n_\text{chan in}$ columns. Each block is a Toeplitz matrix representation of the convolution with a kernel $\theta$. 
\begin{equation}
\TK \equiv 
\begin{bmatrix}
K(\theta^{1,1}) & K(\theta^{1,2}) & \hdots & K(\theta^{1,n_\text{chan in}}) \\ 
K(\theta^{2,1}) & K(\theta^{2,2}) & \hdots & K(\theta^{2,n_\text{chan in}}) \\ 
\vdots & \vdots & \ddots & \vdots \\
 K(\theta^{n_\text{chan out},1}) & K(\theta^{n_\text{chan out},2}) & \hdots & K(\theta^{n_\text{chan out},n_\text{chan in}}) \\ 
\end{bmatrix}.
\end{equation}
This block matrix is square if the number of channels remains the same after the convolutions. To reduce the number of channels, $\TK$ needs to be a flat matrix; a tall block-matrix will increase the number of channels. The collection of convolutional kernels at layer $j$ is denoted by $\Ttheta_j$.

Before introducing the symmetric block-low-rank layer, we recall the basic design of reversible hyperbolic networks \cite{Chang2017Reversible,lensink2019fully}. The foundation of the network is the nonlinear Telegraph equation  \cite{Zhou2018Telegraph},
 \begin{eqnarray}
 \label{telegraph}
 \ddot{\bfY} = f(\bfY,\bftheta(t)),
 \end{eqnarray}
where $ \ddot{\bfY}$ is the second time derivative of the state and the model parameters $\bftheta(t)$ are time dependent. A Leapfrog discretization of the second derivative leads to 
\begin{equation}\label{LeapFrog}
\ddot{\bfY}  \approx {\frac 1 {h^2}} \left(\bfY_{j+1} - 2 \bfY_j + \bfY_{j-1} \right),
 \end{equation}
which is a conservative scheme and $h$ indicates the time step. An important component of networks for image-to-image and video-to-video mappings is coarsening/refining while increasing/decreasing the number of channels. \cite{lensink2019fully} propose to combine these two operations in an invertible manner using orthogonal wavelets, for instance, the Haar transform \cite{WaveletReview}. Other invertible transforms \cite{DinhSB16} may also be used. Including this transform results in the discrete network description
 \begin{eqnarray}
\label{resnethyper}
\bfY_{j+1} = 2 \bfW_j\bfY_j -  \bfW_{j-1} \bfY_{j-1} +  f( \bfW_j \bfY_j,\TK_j),
\end{eqnarray}
where $\TW_j$ is the inverse/forward Haar transform if we change the number of channels and resolution. $\TW_j$ is the identity map we maintain resolution and the number of channels. In the next section, we discuss challenges and solutions for selecting the nonlinearity $f( \bfW_j \bfY_j,\TK_j)$, and why the combination of fully reversible networks and specific choices of $f( \bfW_j \bfY_j,\TK_j)$ result in low memory requirements for network states and convolutional kernels.

\section{Symmetric block-low-rank layers: reducing the number of convolutions kernels while preserving the number of channels}

A widely used neural-network layer is given by 
\begin{equation}
f(\bfK_j \bfW_j \bfY_j + \bfb_j),
 \end{equation} 
where the notation is as in the previous section, and $\bb_j$ indicates a bias term. To make this layer suitable for the fully reversible networks described above, $\bfW_j$ increases the number of channels by a factor eight when we want to coarsen in each of the three spatial/temporal directions of the input data by a factor two, for example, an RGB video. It follows that the number of convolutional kernels is now $64$ times larger than the previous layer, and coarsening a few times leads to tens of thousands of kernels. It is not possible to use a flat convolutional block-matrix $\TK$ that reduces the number of channels because there would be a mismatch with the number of channels in the other terms in the same layer, see equation \eqref{resnethyper}.

However, if we use a symmetric layer \cite{RuthottoHaber2018} of the form
\begin{equation}
f( \bfY_j,\bftheta_j) = -\bfK_j^{\top} f(\bfK_j \bfY_j + \bfb_j),
\end{equation}

we gain some flexibility to design layers for the network with more desirable properties. Our core contribution is using a flat block-convolution matrix $\TK$ in the context of reversible networks with symmetric layers. For example, if there are three channels in $\bfY_j$, we need to output three channels as well, but at the same time want to use only six convolutional kernels instead of the usual nine. We achieve this using a layer with block structure (omitting the layer indicator $j$)

\begin{align}\label{BLRLayer}
&\begin{bmatrix}
K(\Ttheta^{1,1})^\top & K(\Ttheta^{2,1})^\top\\
K(\Ttheta^{1,2})^\top & K(\Ttheta^{2,2})^\top\\
K(\Ttheta^{1,3})^\top & K(\Ttheta^{2,3})^\top
\end{bmatrix} \times \nonumber\\
&f \bigg(
\begin{bmatrix}
K(\Ttheta^{1,1}) & K(\Ttheta^{1,2}) & K(\Ttheta^{1,3})\\
K(\Ttheta^{2,1}) & K(\Ttheta^{2,2}) & K(\Ttheta^{2,3})\\
\end{bmatrix}
\begin{bmatrix}
Y^1 \\
Y^2 \\
Y^3 \\
\end{bmatrix}
+\bfb \bigg).
\end{align}

For simplicity, we assumed that there is no coarsening via the Haar transform at this particular layer. This structure sets a minimum number of convolutional kernels equal to the number of input channels. In the numerical examples, we choose the ReLU function as the nonlinear pointwise activation function $f(\cdot) : \mathbb{R}^N \rightarrow \mathbb{R}^N$. The action of $f(\cdot)$ is equivalent to a diagonal matrix with zero and one on the diagonal, depending on the input. An equivalent point of view is that the ReLU function $f(\cdot)$ sets part of the rows (not block-rows) in $\TK$ to zero. Let us denote $f(\cdot)$ for specific inputs $\{ \TY, \bfb \}$ as the diagonal matrix $\TD$. The symmetric layer then takes the form $- \TK_j^\top \TD_j \TK_j (\TY_j + \bfb_j)$ and $ \TK_j^\top \TD_j \TK_j$ thus remains symmetric-positive-semidefinite for functions $f(\cdot)$ that lead to a diagonal $\TD_j$ with non-negative entries. For this type of function, we can use the entry-wise matrix square-root to write $- \TK_j^\top \sqrt{\TD_j} \sqrt{\TD_j} \TK_j (\TY_j + \bfb_j)$, highlighting the symmetry. Of course, for the ReLU, we have $\TD = \sqrt{\TD}$ but the relation in the previous sentence is valid for general diagonal non-negative $\TD$ and corresponding $f(\cdot)$. The symmetric structure leads to guaranteed stable forward-propagation through the network for an appropriate choice of $h$ \cite{RuthottoHaber2018}.

%\begin{equation}\label{relu_matrix}
%-\bfK_j^{\top} f(\bfK_j \bfY_j + \bfb_j) =
%\begin{bmatrix}
%1 &    &     &  \\
%   & 0 &     &   \\
%   &     & 0 &     \\
%   &     &     &  1\\
%\end{bmatrix}
%\end{equation}

The structure as described above induces several properties, including
\begin{itemize}
\item $\TK$ contains $m \times n$ blocks where we assume $m \leq n$ and the number of convolutional kernels is given by $mn$.
\item the number of convolutions plus transposed convolutions is $2mn$ per symmetric layer. 
\item the block-rank of $\TK^\top \TK$ or $\TK^\top \TD \TK$ is at most $m$.
\item The number of unique kernels in $\TK^\top \TK$ is at most $(n^2 + n)/2$.
\item non-negative point-wise activation functions $f(\cdot)$ preserve symmetry and positive-semidefinitness of $\TK^\top \TK$.
\end{itemize}

A key observation is that a non-square $\TK$ does not affect reversibility of the network. Reverse propagation still adheres
\begin{align}\label{rev_prop}
&\TY_j = \TW_j^{-1} \bigg[ 2 \TW_{j+1} \TY_{j+1} - \nonumber\\
&h^2 \TK_{j+2}^\top  f ( \TK_{j+2} \TW_{j+1} \TY_{j+1} + \bfb_{j+2} ) - \TY_{j+2} \bigg].
\end{align}
Reverse propagation allows the recomputation of network states $\TY_j$ that we need to compute the gradient and avoid storing all $\TY_j$. Now we can also lower the block-rank of $\TK^\top \TK$ to decrease the memory for storing convolutional kernels and reduce the computational cost. Together, these tools extend the applicability of fully reversible hyperbolic networks to larger input data and an increased number of coarsening stages in the network. See Figure \ref{fig:memorycurves} for a summary of the memory requirements.

\section{Selection of the block rank}
Now that we have a strategy to control the memory for the convolutional kernels via the block rank of $\TK^\top \TK$, we concern ourselves with selecting a `good' rank. A simple way is to check how much memory we can spend on convolutional kernels, given the data size and network depth, and adjust the size of $\TK$ at every layer accordingly. This strategy does not exploit the potential benefits of implicit regularization of the symmetric block-low-rank layer, nor does it offer any savings in terms of computational cost.

Our first attempt to minimize computational cost and explore the effects of implicit low-rank regularization is an adaptive approach. Recall that $\TK$ has $m \times n$ blocks (convolutional kernels). We start training with the same maximum block-rank ($m$) for every layer. If we detect that the loss functions stops decreasing, we can add convolutional kernels and increase the block rank without modifying the existing kernels. If we detect overfitting, we can reduce the number of kernels. Note that even if we select $m$ to be the same for all layers, later layers in the network contain more convolutional kernels because $n$ still grows as we coarsen and increase the number of channels via subsequent Haar transforms. The adaptive nature of the symmetric block-low-rank layer makes it easy to adapt without modifying any software.

\section{Numerical examples}
Two numerical examples highlight the primary contributions of this work. We show an example of a fully reversible hyperbolic network with three coarsening stages, see equations \eqref{resnethyper} and \eqref{BLRLayer}. Training this network (Table \ref{tab:NetworkDesign}, third column) using a `standard' layer structure as in previously proposed reversible networks would not be feasible using GPUs that have around 12GB of memory. Table \ref{tab:NetworkMem} lists the memory requirements for a regular fully reversible hyperbolic network and for versions with symmetric block-low-rank layers. The data size for this experiment is $240 \times 424 \times 72 \times 6$; ($2\times$ RGB channels) .

\begin{table}[]
\begin{center}
\begin{tabular}{|l|ll|}
\hline
Layer  &  Stage 1 &  Stage 2  \\
\hline\hline
1-3 &  $ 6 \times 6$ & $6 \times 6$ \\
4-7 &  $ 4 \times 48$ & $8 \times 48$ \\
7-11  & $4 \times 384$ & $8 \times 384$ \\
12-15 &  $ 4 \times 3072$ & $ 8 \times 3072$ \\
16-19 &  $4 \times 384$ & $8 \times 384$ \\
20-23 &  $4 \times 48$ & $8 \times 48$ \\
24-27 &  $6 \times 6$ & $6 \times 6$\\
\hline
\end{tabular}
\end{center}
\caption{Number of convolutional kernels in (shape of the block matrix) $\TK$ at every layer in the fully reversible network (\eqref{resnethyper}, \eqref{BLRLayer}) with three coarsening steps, for each of the two training stages for our numerical examples.}
\label{tab:NetworkDesign}
\end{table}

\begin{table}[]
\begin{center}
\begin{tabular}{|l|lll|}
\hline
Memory in MB. & BLR=4  & BLR=8 &  Full   \\
\hline\hline
Mem. conv. kernels &7 & 14 & 4206  \\
Mem. states & 528   & 528  &  528  \\
Mem. total & 534 & 541 &  4734\\
\hline
\end{tabular}
\end{center}
\caption{Memory requirements for the network states and convolutional kernels for a fully reversible network as in Table \ref{tab:NetworkDesign}. Shown for the regular fully reversible hyperbolic network (full), and for the version proposed in this work that uses symmetric block-low-rank (BLR) layers.}
\label{tab:NetworkMem}
\end{table}

\begin{figure}
\begin{center}
   \includegraphics[width=1.0\linewidth]
                   {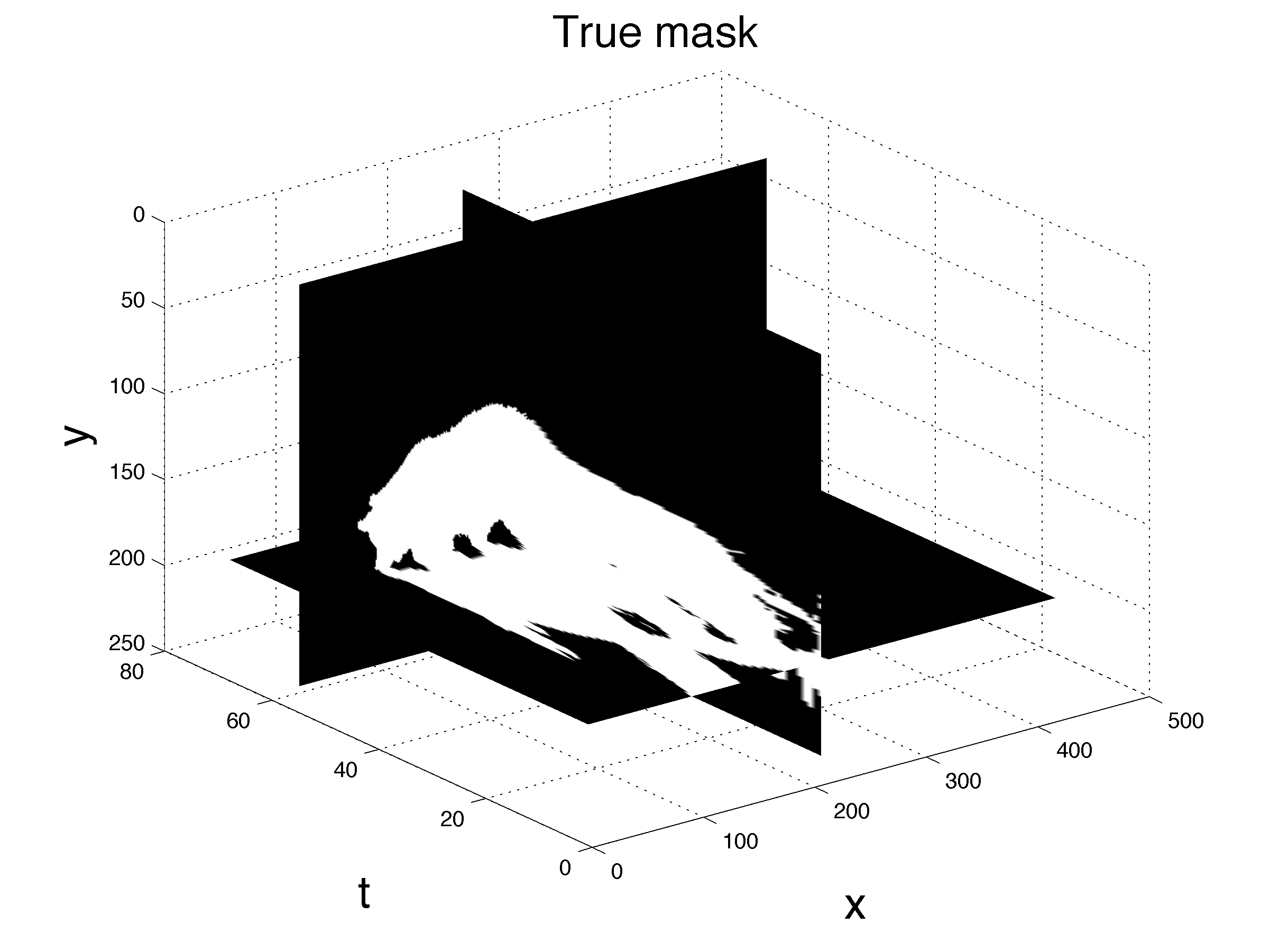}
\end{center}
\caption{True label for the RGB video segmentation problem.}
\label{fig:TruelBear}
\end{figure}

\begin{figure}
\begin{center}
   \includegraphics[width=1.0\linewidth]
                   {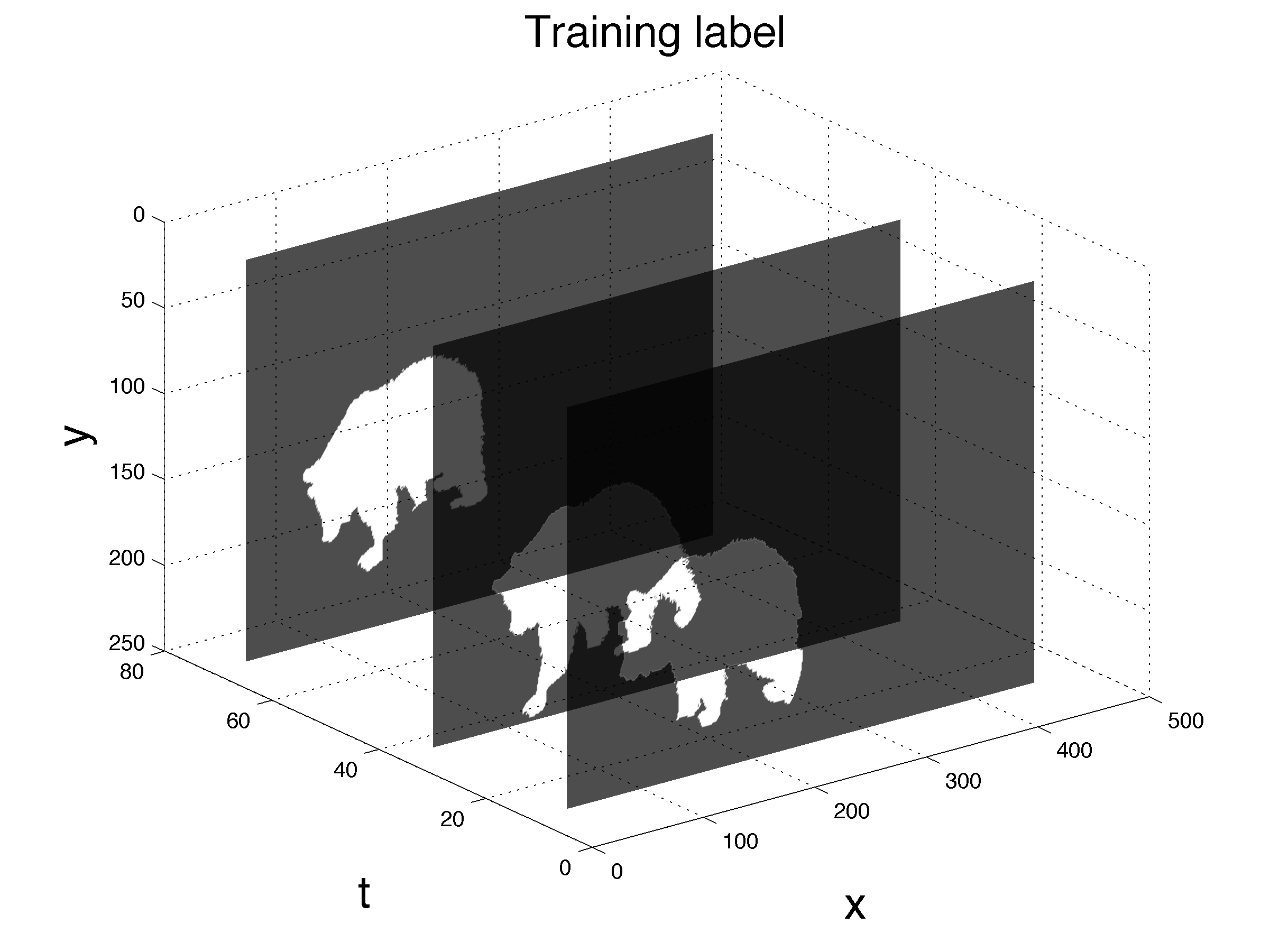}
\end{center}
\caption{Training label for the RGB video segmentation problem: there are three time-slices available for training. There are no other videos in the training set for this task}
\label{fig:LabelBear}
\end{figure}

The first example is an adaptation from \cite{lensink2019fully}. The goal is to segment a video by training a network to map the entire video to its segmentation (Figure \ref{fig:TruelBear}) directly. The specific task is to train on a single RGB video from the Davis video dataset \cite{IEEEDavisDataset}, where only three time-slices have annotations (Figure \ref{fig:LabelBear}). Rather than aiming to generalize the segmentation power to new videos, this experiment tests the ability to work with few data and even fewer labels. Real-time inference is not the aim of this experiment and similar tasks that occur in medical \cite{Unet3D} and geophysical data interpretation \cite{doi:10.1190/INT-2018-0225.1}. 

\begin{figure}
\begin{center}
   \includegraphics[width=1.0\linewidth]
                   {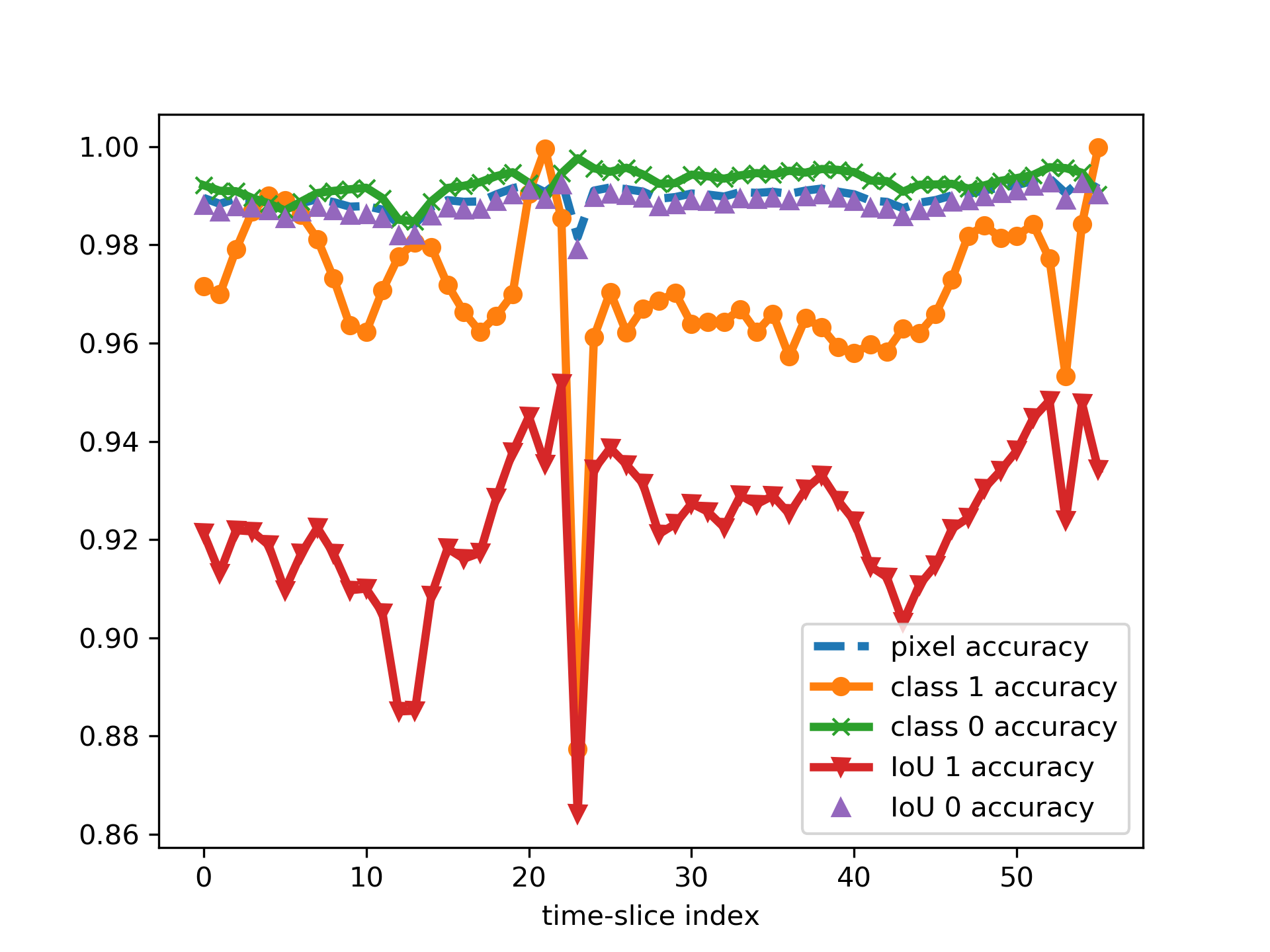}
\end{center}
\caption{Prediction accuracy per time-slice for the RGB video segmentation example.}
\label{fig:SliceIoU}
\end{figure}

\begin{figure}
\begin{center}
   \includegraphics[width=1.0\linewidth]
                   {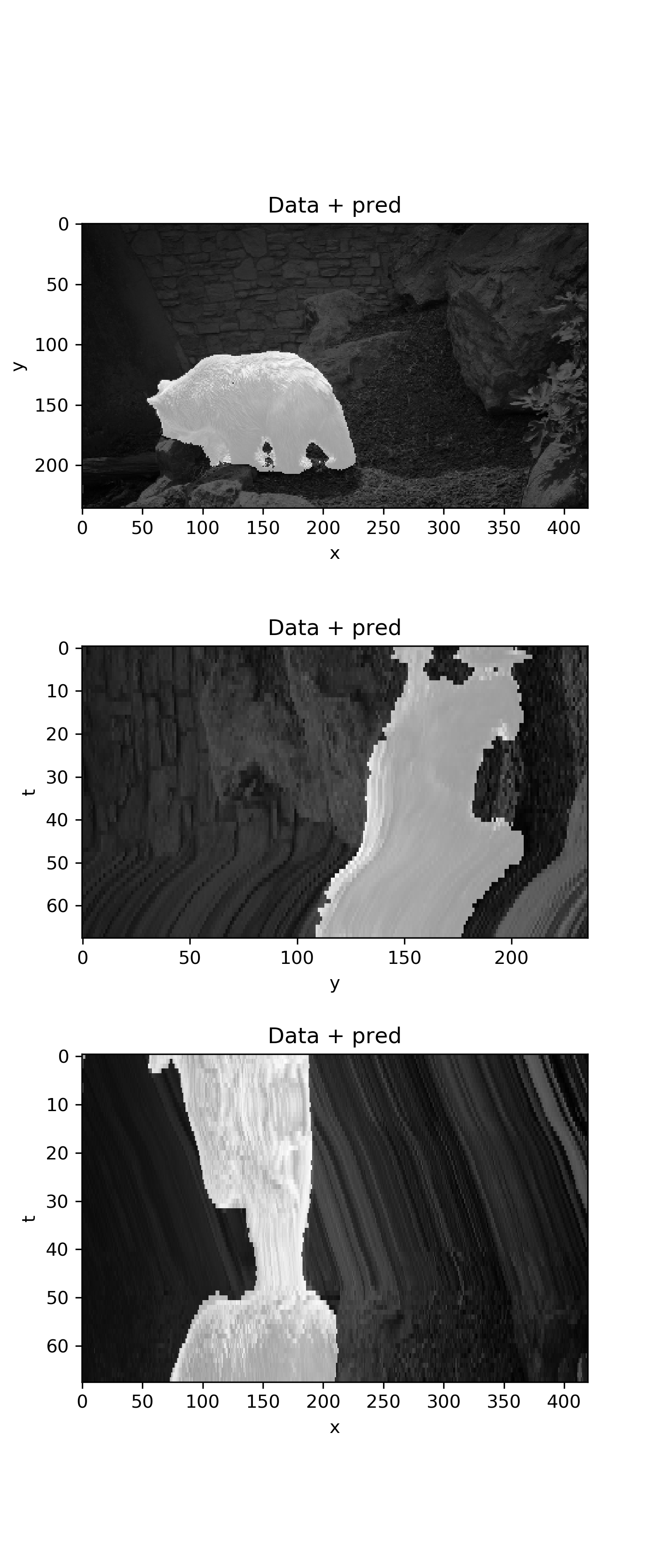}
\end{center}
\caption{Three orthogonal slices from the prediction (white), overlaid on the RGB data (shown in grayscale).}
\label{fig:PredictionBear}
\end{figure}

Different from \cite{lensink2019fully}, we employ layers with the symmetric block-low-rank structure to reduce the number of network parameters and compute much fewer convolutions per gradient computation. Table \ref{tab:Results} shows the results using a limited block-rank of $8$. This table also shows that if we first train with a maximum block rank of $4$, we can add convolutional kernels when the loss-function decrease slows down, and achieve similar good results. Figure \ref{fig:PredictionBear} shows slices of the prediction, and Figure \ref{fig:SliceIoU} shows that the results are accurate across all time-slices.

\begin{table}[]
\begin{center}
\begin{tabular}{|l|cc|}
\hline
 Video segmentation & $\text{BLR}=4 \rightarrow 8$ & $\text{BLR}=8$  \\
\hline\hline
Mean IoU. & $95.5 \%$ & $ 94.5 \%$\\
\hline
\end{tabular}
\end{center}
\caption{Results in terms of mean Intersection over Union (IoU) for the video segmentation example. The column $\text{BLR}=8$ used a fixed maximum block-rank of $8$. $\text{BLR}=4 \rightarrow 8$ indicates that training started with a maximum block-rank of $4$, followed by more training using a maximum block-rank of $8$. Both methods use the same number of iterations in total. }
\label{tab:Results}
\end{table}

\section{Discussion}
We proposed a solution to the problem of the extremely rapid growth of the number of convolutional kernels in fully reversible hyperbolic networks with multiple invertible coarsening layers. A secondary benefit is that the specific use of symmetric block-low-rank network layers provides a way to control the computational cost for each layer while maintaining a possibly large number of channels. This layer structure is general and fits in any network that has layers for which the number of input and output channels is the same. Therefore, we can replace the symmetric block-low-rank layer with other proposed layers that reduce the memory for network parameters. We compare a few of such methods below.

\cite{ephrath2019leanresnet} propose a $\TK$ that contains convolutional kernels on its block-diagonal and scalars ($1 \times 1 \times 1$ convolutions) on every off-diagonal element. The memory for $\TK$ for one layer is then proportional to $3 \times 3 \times 3 \times n_\text{chan} + (n_\text{chan}^2 - n_\text{chan}) \times 1$, which is further reduced if we employ seven-point stencils rather than a $27$-point stencil \cite{ephrath2019leanresnet}. Our proposed block-low-rank layer has memory requirements that scale as $m n_\text{chan} \times 3 \times 3 \times 3$, where $m$ is the selected rank, so we do not depend quadratically on the number of channels. Another main difference is that our block low-rank layer is effectively filled with full convolutions via $\TK^\top \TK$, as opposed to layers that contain mostly scalars.

\cite{8686678,treister2018low} use a block-circulant structured $\TK$. Naive computation of $\TK \TY$ would be unnecessarily expensive, and an FFT-based evaluation of this quantity is required. A bock-circulant $\TK$ has $n_\text{chan}$ unique convolutional kernels. A limitation is that block-circulant is a fixed structure, and we cannot trivially extend the number of kernels to obtain a more expressive $\TK$.

\cite{Szegedy_2015_CVPR,iandola2016squeezenet} present squeeze-expand bottleneck network layers that are related to ours. In particular, \cite{iandola2016squeezenet} use a structure that is equivalent to $f(\TK_2 f(\TK_1 \TY_1))$, where $\TK_1$ is a flat block-matrix with $1 \times 1$ convolutional kernels, and $\TK_2$ is a tall block-matrix with a mix of $1 \times 1$ and $3 \times 3$ convolutional kernels. Besides the kernel sizes and the additional nonlinearity (ReLU in their case), also note that $\TK_1 \TK_2$ is generally not positive-semidefinite; something we do guarantee by construction of the symmetric layer.

While a numerical comparison is beyond the scope of this work, it is clear there are multiple contenders to achieve memory savings in fully reversible hyperbolic networks. As shown in this work, we propose to use the symmetric block-low-rank layer because it does not need special matrix-vector product implementations or other software modifications; we can easily add or remove convolutional kernels while training; and the use of matrix-vector product notation and the low-rank interpretation allows us to observe a number of useful properties of the proposed layer.

\section{Conclusions}
We proposed a symmetric network layer with a low block-rank for fully reversible network for problems with large data inputs like video-to-video mappings. While fully reversible networks avoid storing all network states to compute a gradient of the loss functions, the dominant memory factor becomes the storage of convolutional kernels. This is especially limiting for networks that contain multiple invertible coarsening operations that also change the number of channels. The symmetric block-low-rank layers allow us to train the network in a factorized form directly. It also provides the freedom to select the desired memory usage for convolutional kernels without requiring any changes to the network design or code. Besides enabling large data inputs for a fully reversible network with multiple coarsening steps, we also propose an adaptive-rank training strategy based on the same layer. Training starts with a low block-rank for every layer, followed by further training while increasing/decreasing the block-rank by augmenting/removing convolutional kernels to the already trained ones.

Numerical experiments show that we can train a fully reversible network on video input with good results. This experiment would not have been possible with a regular fully reversible network. Moreover, we also show that the incremental rank training strategy provides similar results. While the numerical experiments only illustrate the material presented, the ability to train from video to video or 3D medial/geophysical imaging to 3D interpretation simplifies existing approaches that operate on 2D slices or small 3D sub-volumes.

{\small
\bibliographystyle{ieee_fullname}
\bibliography{biblio,VideoRefs}
}

\end{document}